\PassOptionsToPackage{table,xcdraw}{xcolor}
\documentclass[sigconf]{acmart}
\usepackage{enumitem}
\usepackage{multirow,multicol}
\AtBeginDocument{%
\providecommand\BibTeX{{%
\normalfont B\kern-0.5em{\scshape i\kern-0.25em b}\kern-0.8em\TeX}}}

\setcopyright{acmlicensed}
\copyrightyear{2025}
\acmYear{2025}
\acmDOI{XXXXXXX.XXXXXXX}
\acmConference[KDD '25]{Make sure to enter the correct
  conference title from your rights confirmation emai}{August 03--07,
  2025}{Toronto, ON, Canada}
\acmISBN{978-1-4503-XXXX-X/18/06}

\begin{document}

\title{BatteryLife: A Comprehensive Dataset and Benchmark for Battery Life Prediction}


\author{Ruifeng Tan}
\authornote{Authors contributed equally to this research.}
\author{Weixiang Hong}
\authornotemark[1]
\author{Jiayue Tang}
\author{Xibin Lu}
\affiliation{%
	\institution{The Hong Kong University of Science
		and Technology (Guangzhou)}
	\city{Guangzhou}
	\state{Guangdong}
	\country{China}
}
\affiliation{
	\institution{The Hong Kong University of Science
		and Technology}
	\city{Hong Kong SAR}
	\country{China}
}
\email{rtan474@connect.hkust-gz.edu.cn}
\email{whong719@connect.hkust-gz.edu.cn}
\email{jtang638@connect.hkust-gz.edu.cn}
\email{xlubc@connect.hkust-gz.edu.cn}

\author{Ruijun Ma}
\author{Xiang Zheng}
\affiliation{%
	\institution{CALB Group Co., Ltd.}
	\city{Changzhou}
	\country{China}}
\email{ruijun.ma@calb-tech.com}
\email{xiang.zheng@calb-tech.com}

\author{Jia Li}
\authornote{Corresponding authors}
\author{Jiaqiang Huang}
\authornotemark[2]
\author{Tong-Yi Zhang}
\authornotemark[2]
\affiliation{%
	\institution{The Hong Kong University of Science
		and Technology (Guangzhou)}
	\city{Guangzhou}
	\state{Guangdong}
	\country{China}
}
\affiliation{
	\institution{The Hong Kong University of Science
		and Technology}
	\city{Hong Kong SAR}
	\country{China}
}
\email{jialee@ust.hk}
\email{seejhuang@hkust-gz.edu.cn}
\email{mezhangt@hkust-gz.edu.cn}

\renewcommand{\shortauthors}{Ruifeng Tan et al.}

\begin{abstract}
	Battery Life Prediction (BLP), which relies on time series data produced by battery degradation tests, is crucial for battery utilization, optimization, and production. Despite impressive advancements, this research area faces three key challenges. Firstly, the limited size of existing datasets impedes insights into modern battery life data. Secondly, most datasets are restricted to small-capacity lithium-ion batteries tested under a narrow range of diversity in labs, raising concerns about the generalizability of findings. Thirdly, inconsistent and limited benchmarks across studies obscure the effectiveness of baselines and leave it unclear if models popular in other time series fields are effective for BLP. To address these challenges, we propose BatteryLife, a comprehensive dataset and benchmark for BLP. BatteryLife integrates 16 datasets, offering a 2.5 times sample size compared to the previous largest dataset, and provides the most diverse battery life resource with batteries from 8 formats, 59 chemical systems, 9 operating temperatures, and 421 charge/discharge protocols, including both laboratory and industrial tests. Notably, BatteryLife is the first to release battery life datasets of zinc-ion batteries, sodium-ion batteries, and industry-tested large-capacity lithium-ion batteries. With the comprehensive dataset, we revisit the effectiveness of baselines popular in this and other time series fields. Furthermore, we propose CyclePatch, a plug-in technique that can be employed in various neural networks. Extensive benchmarking of 18 methods reveals that models popular in other time series fields can be unsuitable for BLP, and CyclePatch consistently improves model performance establishing state-of-the-art benchmarks. Moreover, BatteryLife evaluates model performance across aging conditions and domains. BatteryLife is available at \url{https://github.com/Ruifeng-Tan/BatteryLife}.
\end{abstract}


\maketitle

\section{INTRODUCTION}
Rechargeable batteries are ubiquitous in modern industry, including electric vehicles, power grids, and portable devices \cite{20221311845974, tao2023collaborative, LU2022100207}. Nevertheless, batteries inevitably degrade with cyclic operation due to intrinsic electrochemical mechanisms \cite{BIRKL2017373, 20150100399520, TAN2024103725}. The degradation causes shorter usable time and may lead to safety issues in the applications \cite{20243216836174, zhang2023realistic}. Typically, batteries are deemed at end-of-life when their capacity falls to 80\% of the initial, and end-of-life batteries are unsuitable for high-demanding applications like electric vehicles \cite{20191406719359, attia2020closed, 20223412623287}. To ensure more sustainable and safer battery operation, battery engineers/scientists need to conduct degradation tests to measure the battery life during battery optimization and quality control \cite{attia2020closed, LI2024101891, STOCK2022104144}. However, degradation tests are time-consuming since non-linear capacity loss occurs over months to years before the end of life. Accurate early prediction of the battery life is thus vital for battery utilization, optimization, and production.

Machine learning models have been developed as promising approaches for Battery Life Prediction (BLP) \cite{20191406719359,HU2020310,JIANG20213187, 20214611166935, 20223412623287, ZHOU2024113749, HSU2022118134, 8289406, FEI2023106903,zhang2025battery,9137406,tao2023battery}. In early works, battery experts generally made arduous efforts to extract features from the voltage as well as current time series of degradation tests \cite{20191406719359,JIANG20213187} and data of additional tests such as temperature test \cite{zhang2025unlocking} and acoustic test \cite{20181705037599} based on domain knowledge. The extracted features are then fed into classic machine learning models such as Gaussian process regressor to predict the battery life. Recently, with the advent of larger datasets, neural networks have shown advantages over classic machine learning models without requiring deep domain expertise \cite{20223412623287, XU2021107396, TANG2022111530, FEI2023106903,zhang2025battery,tao2023battery}. Despite the success these methods have achieved, they also reveal three issues that impede advances in this field:

\begin{table*}[!t]
	\caption{Data statistics of battery life datasets. In this paper, a chemical system is a combination of cathode, anode, and electrolyte, and a protocol indicates one set of charge and discharge configurations adopted during cycling.} \label{DatasetStatistics}
	\resizebox{1.0\textwidth}{!}{
		\begin{tabular}{lllllllll}
			\bottomrule
			Dataset  & Battery Number & Format & Chemical System & Operation Temperature & Protocol & Data Source               & Battery Type           & Year \\
			\hline
			CALCE \cite{20132816492471, HE201110314}     & 13             & 1                 & 1               & 1                     & 2                 & Lab test                  & Li-ion                 & 2013     \\
			HNEI \cite{20182105222726}      & 14            & 1                 & 1               & 1                     & 2                 & Lab test                  & Li-ion                 & 2018     \\
			MATR \cite{attia2020closed, 20191406719359}      & 169            & 1                 & 1               & 1                     & 81                & Lab test       & Li-ion        & 2019,2020     \\
			UL\_PUR \cite{juarez2020degradation, Juarez-Robles_2021}   & 10            & 1                 & 1               & 1               & 1                 & Lab test                  & Li-ion                 & 2020     \\
			SNL \cite{20203809192263}      & 50             & 1                 & 3               & 3                     & 23                & Lab test                  & Li-ion                 & 2020     \\
			MICH\_EXP \cite{mohtat2021reversible}     & 18             & 1                 & 1               & 3                     & 6                & Lab test                  & Li-ion                 & 2021     \\
			MICH \cite{20214611166935}     & 40           & 1                 & 1               & 2                     & 2                & Lab test                  & Li-ion                 & 2021     \\
			RWTH  \cite{LI2021230024}   & 48           & 1                 & 1               & 1                     & 1                 & Lab test                  & Li-ion                 & 2021     \\
			HUST \cite{20223412623287}      & 77             & 1                 & 1               & 1                     & 77                & Lab test                  & Li-ion                 &  2022    \\
			Tongji \cite{zhu2022data}    & 108          & 1                 & 3               &   3                    & 6                & Lab test                  & Li-ion                 &    2022  \\
			BatteryArchive \cite{batteryarchive} & 139 & 3 & 8 & 6 & 36 & Lab test & Li-ion & 2024\\
			Stanford \cite{20244517308214} & 41            & 1                 & 1               & 1                     & 1                & Lab test                  & Li-ion                 & 2024     \\
			XJTU \cite{wang2024physics}      & 23             & 1                 & 1               & 1                     & 2                 & Lab test                  & Li-ion                 & 2024     \\
			ISU\_ILCC \cite{LI2024101891} & 240           & 1                 & 1               & 1                     & 202               & Lab test                  & Li-ion                 & 2024     \\
			BatteryML \cite{20243216836174} & 381           & 2                 & 5               & 5                     & 186               & Lab test                  & Li-ion                 & 2024     \\
			BatteryLife      & 990          & 8          & 59         & 9          & 421               & Lab test; Industrial test & Li-ion; Zn-ion; Na-ion &  2025  \\
			\toprule
		\end{tabular}
	}
\end{table*}

\begin{itemize}[left=0pt]
	\item \textbf{Limited dataset size.} As shown in Table \ref{DatasetStatistics}, the previous largest dataset is BatteryML \cite{20243216836174} that offers 381 end-of-life batteries by integrating 7 public datasets \cite{attia2020closed, 20191406719359, 20223412623287, 20182105222726, 20203809192263, juarez2020degradation, LI2021230024, Juarez-Robles_2021}. However, this undertaking overlooks several datasets \cite{zhu2022data,20214611166935,mohtat2021reversible} and lacks datasets \cite{20244517308214,wang2024physics,LI2024101891} that are released recently. In this vein, BatteryML, whose size is smaller than half of the sum of existing datasets, fails to provide comprehensive insights into modern battery life data.
	\item  \textbf{Limited data diversity.} According to Table \ref{DatasetStatistics}, most existing battery life datasets offer only lithium-ion (Li-ion) batteries with one format, one chemical system, one operation temperature, and several charge/discharge protocols. The BatteryML \cite{20243216836174} and ISU\_ILCC \cite{LI2024101891} offer great diversity in charge/discharge protocols, but their diversity in other aspects is still limited. Researchers \cite{20223412623287} have observed that models validated on various charge protocols may be ineffective in learning varying discharge protocols. This raises concerns about the generalizability of findings derived from datasets with limited diversity. Furthermore, the limited diversity results in a lack of data needed to inform broader battery research directions.
	\item \textbf{Inconsistent and limited benchmarks.} Existing works \cite{20191406719359,HU2020310,JIANG20213187, 20214611166935, 20223412623287, ZHOU2024113749, HSU2022118134, 8289406, FEI2023106903,zhang2025battery} generally evaluate models using different datasets and varied experimental setups (e.g. different baselines and data preprocessing). Additionally, these studies utilize data spanning different numbers of cycles to predict battery life, ranging from a single cycle \cite{HSU2022118134} to one hundred cycles \cite{20191406719359,zhang2025battery}, leading to inconsistent task settings. Moreover, current benchmarks often adopt a limited set of classic baselines (e.g. vanilla recurrent neural networks). This lack of a consistent and comprehensive benchmark complicates the evaluation of baselines in this field and leaves it uncertain whether models that are popular in other time series fields (e.g. weather, traffic, and electricity) are effective for BLP.
\end{itemize} 

To address these issues, we propose BatteryLife, a comprehensive dataset and benchmark for BLP. From a dataset perspective, BatteryLife stands out as the largest and most diverse battery life dataset by integrating 16 datasets. Specifically, BatteryLife is 2.5 times the size of BatteryML \cite{20243216836174} in terms of battery number, covering 90,000 samples. Remarkably, BatteryLife furnishes unparalleled diversity, delivering 4 times format, 11.8 times chemical system, 1.8 times operation temperature, and 2.2 times charge/discharge protocol compared to BatteryML \cite{20243216836174}. Notably, BatteryLife pioneers as the first work that releases zinc-ion (Zn-ion) and sodium-ion (Na-ion) battery life datasets, and additionally offers large-capacity Li-ion batteries that are tested by the battery manufacturer (i.e., CALB Group Co., Ltd.). The formidable diversity challenges the model's capability to learn essential mappings from early data to life and allows inspiration for the data mining community and a broader battery community.

\begin{figure}[t]
	\centering
	\includegraphics[width=0.95\linewidth]{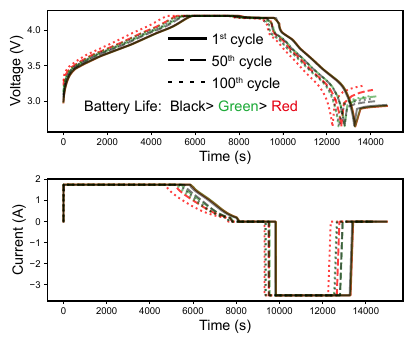}
	\caption{Partial cycling data (voltage and current time series) from three batteries with the same specifications tested under a protocol.}
	\Description[zero figure]{Partial cycling data (voltage and current time series) from three batteries with the same specifications tested under a protocol.}
	\label{demoFigure}
\end{figure}

With respect to benchmark, BatteryLife provides a unified experimental setup and supports BLP across cycles ranging from one to one hundred, covering a wide range of application scenarios. Moreover, BatteryLife serves as a comprehensive benchmark for BLP. It encompasses methods popular in this and other time series fields. Additionally, motivated by the patching technique \cite{DBLP:conf/iclr/NieNSK23} and the inherent characteristics of degradation tests, we propose CyclePatch, a plug-in technique for modeling degradation test data. As illustrated in Figure~\ref{demoFigure}, for batteries with the same specifications, voltage and current time series exhibit similar patterns across cycles within a protocol. This observation motivates us to treat each cycle as a token in CyclePatch, explicitly capturing recurring patterns in degradation tests. Furthermore, Figure~\ref{demoFigure} demonstrates that cycling data from batteries with different life labels show discriminative characteristics, both in terms of cycles of the same number and variance across cycles. Thus, CyclePatch employs an intra-cycle encoder to model the complex interactions among variables within each cycle. In this framework, CyclePatch generates informed representations for each cycle, on which an inter-cycle encoder is applied to learn effective patterns across cycle tokens. Through extensive experiments of 18 methods, we reveal that some techniques widely adopted in other time series fields are inappropriate for BLP. Additionally, models consistently achieve improved performance when CyclePatch is integrated. Moreover, the intricacies of achieving effective BLP are further disclosed via evaluations across aging conditions and domains.

Our main contributions are summarized as follows:

\begin{itemize}[left=0pt]
	\item  \textbf{The largest battery life dataset:} BatteryLife offers 90,000 samples from 990 batteries, which is 2.5 times the size of the previous largest one. The unprecedented dataset size allows researchers to generate practical insights into battery life data by making advances on BatteryLife.
	\item  \textbf{The most diverse battery life dataset:} BatteryLife contributes the most diverse dataset with considerable increases in all diversity aspects as mentioned above. BatteryLife also presents the first dataset that contains lab-tested Li-ion, Na-ion, and Zn-ion batteries as well as industry-tested large-capacity Li-ion batteries. The unparalleled diversity of BatteryLife requires models that are capable of learning fundamental patterns for BLP. Moreover, the diversity allows advances in BatteryLife to inform the data mining community and a broader battery community.
	\item  \textbf{A comprehensive benchmark for BLP:} BatteryLife is a comprehensive benchmark that provides fair comparisons of a series of popular baselines. In addition, BatteryLife introduces CyclePatch that is a plug-in technique for BLP. By benchmarking 18 models, we offer valuable insights into their effectiveness, shedding light on the intricacies of achieving accurate early BLP.
\end{itemize}

\section{PRELIMINARIES}
\subsection{Degradation Test}\label{SecDegradationTest}
\begin{figure}[t]
	\centering
	\includegraphics[width=1.0\linewidth]{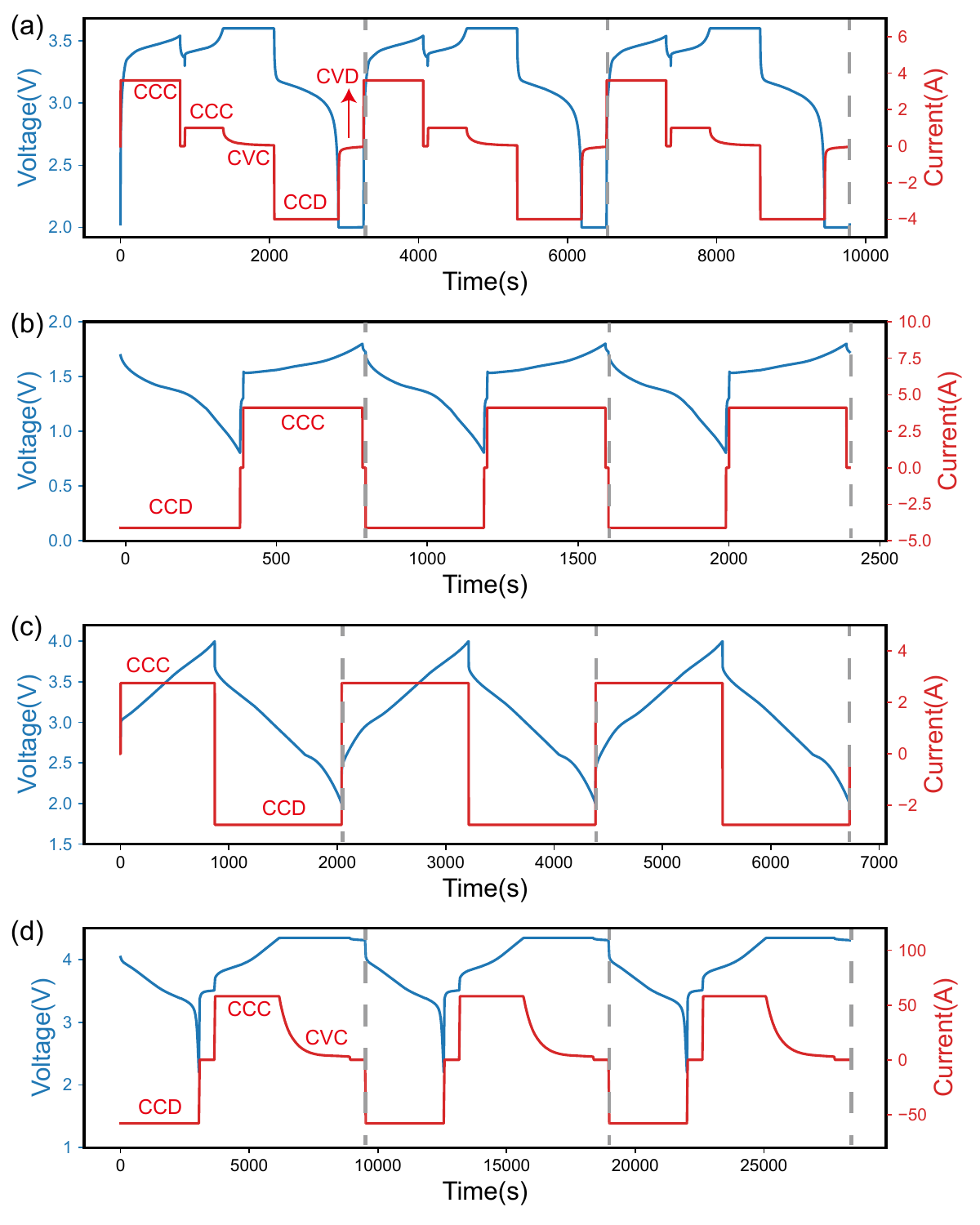}
	\caption{Partial cycling data in degradation tests from (a) a lab-tested Li-ion battery, (b) a lab-tested Zn-ion battery, (c) a lab-tested Na-ion battery, and (d) an industry-tested Li-ion battery. The positive and negative values in current represent charging and discharging, respectively. The gray dotted lines are separators of different cycles. The charging and discharging periods are labeled in the first cycle as illustrations.}
	\Description[first figure]{Partial cycling data in degradation tests from (a) a lab-tested Li-ion battery, (b) a lab-tested Zn-ion battery, (c) a lab-tested Na-ion battery, and (d) an industry-tested Li-ion battery. The positive and negative values in current represent charging and discharging, respectively. The gray dotted lines are separators of different cycles. The charging and discharging periods are labeled in the first cycle as illustrations.}
	\label{BatteryCurve}
\end{figure}

\begin{figure}[t]
	\centering
	\includegraphics[width=0.85\linewidth]{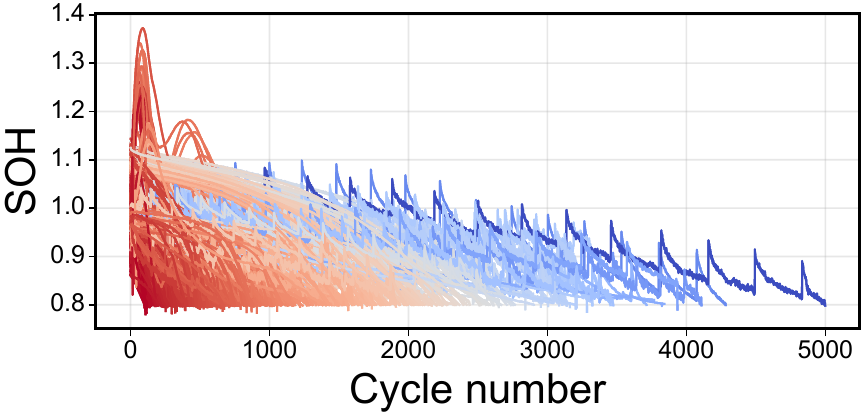}
	\caption{SOH degradation trajectories in BatteryLife.}
	\Description[SOH trajectories]{SOH degradation trajectories in BatteryLife.}
	\label{FigSOHCurve}
\end{figure}

Traditionally, battery life is measured via degradation tests consisting of repeated cycles, where one cycle is composed of a charging and discharging period. The degradation tests are performed following pre-set charge and discharge protocols in temperature-controlled environmental chambers, whereas the protocols and operation temperature are set according to specific interests. In Figure~\ref{BatteryCurve}, we show partial cycling data of four batteries. It can be observed that cycling data are time series, in which fundamental signals are voltage ($V$) and current ($I$). It should be noted that, for any time step, either V or I is controlled by preset protocols, while the uncontrolled one (I or V) can respond to the battery internal state. The capacity is computed as 
\begin{equation}
	Q_{i}=\int_{t_1}^{t_2}\left|I\right|dt.
\end{equation}
By using this equation, we can compute the charge or discharge capacity within a time region for a cycle. When $t_1$ and $t_2$ are the start and end time of a charging or discharging period, the $Q_i$ is the charge or discharge capacity for the $i^{th}$ cycle, respectively. The state of health (SOH) of a battery at $i^{th}$ cycle is calculated by:
\begin{equation}
	\label{EqSOH}
	SOH=\frac{Q_i}{Q_0},
\end{equation}
where $Q_i$ is the capacity of $i^{th}$ cycle, and $Q_0$ is the initial capacity. The literature \cite{20191406719359,20223412623287,20243216836174,wang2024physics,20211510188187,lu2023deep} typically takes discharge capacity as $Q_i$. Regarding  $Q_0$, some works \cite{20191406719359,20223412623287,20243216836174,wang2024physics} utilize nominal capacity as $Q_0$ while some \cite{20211510188187,lu2023deep} set $Q_1$ as $Q_0$. Non-linear SOH degradation can occur hundreds of cycles before batteries reach the end of life (80\% SOH) as shown in Figure~\ref{FigSOHCurve}.

It should be noted that the cycling data pattern is affected by aging conditions \cite{zhang2025battery}. In this work, battery format, anode, cathode, electrolyte, charge protocols, discharge protocols, operation temperature, nominal capacity, and manufacturer are considered aging factors, and a difference in any aging factor generates a different aging condition. To illustrate, vivid examples of the impact of charge protocols and discharge protocols as well as chemical systems are shown in Figure~\ref{BatteryCurve}. Figure~\ref{BatteryCurve}a plots a Li-ion example. The Li-ion example has two-stage constant-current charging (CCC), where different currents are employed in different stages, followed by a constant-voltage charging (CVC) during the charging period. Then the battery undergoes a constant-current discharge (CCD) and constant-voltage discharge (CVD). However, other examples comply with one-stage CCC, and some have no CVC or CVD periods. From Figure~\ref{BatteryCurve}d, it is also observed that the industrial test applies a current that is times larger than the current of laboratory tests (Figure~\ref{BatteryCurve}a-c). Moreover, the battery data pattern is influenced by underlying chemical systems as witnessed in the different shapes of voltage curves in Figure~\ref{BatteryCurve}a-c.  

\subsection{Task Definition}
Let $\textbf{X}_{i:N}=[\textbf{X}_i,\textbf{X}_{i+1},\cdots,\textbf{X}_N]\in\mathbb{R}^{3\times T}$ represent the voltage, current and capacity variables across $T$ time steps starting from the $i^{th}$ cycle to the $N^{th}$ cycle, where $\textbf{X}_i\in\mathbb{R}^{3\times T_i}$ is the cycling data of the $i^{th}$ cycle with $T_i$ time steps. We consider the problem: Given the input $\textbf{X}_{1:S}$ with $\forall S\leq 100$, where $S$ is a positive integer, a model needs to predict the battery life denoted by $y\in\mathbb{R}^{1}$. Unless otherwise specified, we use nominal capacity as $Q_0$ to compute SOH according to Equation~\ref{EqSOH}, and consider the cycle number at which the SOH becomes no larger than 80\% as the battery life following the seminal work of Severson et al. \cite{20191406719359}.

\section{BATTERYLIFE}
\subsection{Dataset Overview}
BatteryLife dataset is constructed via integration of 3 our datasets and 13 previously available datasets \cite{20132816492471,20182105222726,attia2020closed, 20191406719359,juarez2020degradation,20203809192263,LI2021230024,20214611166935, mohtat2021reversible,20223412623287,zhu2022data,20244517308214,wang2024physics,LI2024101891}. The data statistics of BatteryLife and previous datasets are concluded in Table~\ref{DatasetStatistics}. According to Table~\ref{DatasetStatistics}, the previous public datasets contain only Li-ion batteries collected from laboratory tests. We supply Zn-ion and Na-ion batteries by releasing batteries tested in our lab and provide industry-tested large-capacity Li-ion batteries accessed from CALB Group Co., Ltd.. On this note, BatteryLife features as the largest and most diverse battery life dataset. Note that statistics for all datasets are conducted after data preprocessing. We refer readers to Appendix~\ref{AppendixA} for the details of data preprocessing, simple introductions to each dataset, and degradation test details of 3 our datasets. For convenient usage, we standardize all datasets in a pickle file format with unified naming rules following BatteryML \cite{20243216836174}, whereas the original data come from various naming rules and file formats.

We split BatteryLife into four parts: Li-ion (combination of 13 public datasets), Zn-ion, Na-ion, and CALB, and summarize the data statistics of each part in Table~\ref{TStatisticsEach}. Such a division is employed for two reasons. On the one hand, different battery type exhibits different cycling patterns due to intrinsic electrochemical mechanisms, and thus Li-ion, Zn-ion, and Na-ion batteries are categorized as different parts. Likewise, CALB batteries are separated from others because they are large-capacity batteries actively used in electric vehicles and are subject to industrial tests. Therefore, this division separates battery data according to domains and can examine model performance on different domains. On the other hand, the four parts generally intrigue different end users (battery practitioners), and hence allow model advances to inform various users from the battery community.

Collectively, BatteryLife offers standardized data with unprecedented public dataset size and diversity for BLP. This enables researchers to explore lots of neural network design and generate actionable insights by making advances on BatteryLife. 

\subsection{Popular Benchmark Methods}\label{SecBaselines}
We set up a series of benchmark methods in BatteryLife. First, we select methods that are widely adopted for BLP \cite{FEI2023106903,HSU2022118134,20223412623287,zhang2025battery,9137406,WOS:000610934400001}, including Transformer encoder \cite{20182105216120}, Long short-term memory (LSTM), bidirectional long short-term memory (BiLSTM), gate recurrent unit (GRU), bidirectional gate recurrent unit (BiGRU), convolutional neural network (CNN), and multilayer perceptron (MLP). Among these models, Transformer encoder, LSTM, BiLSTM, GRU, and BiGRU are directly applied to process input that has a variable length, and then a linear projection is used to produce the prediction. For CNN, we pad the input to the same length using zeros and then reshape the input into an image-like matrix using voltage, current, and capacity as three channels. Specifically, the reshaped input has a dimension of (3, 100, 300), where 100 is the largest input cycle number, and 300 is the number of resampled points within a cycle. With the reshaped input, 2-dimension CNN can be used to model cycling data and a linear projection is employed to produce predictions based on the flattened output of the CNN. As for MLP, we utilize it to model the flattened zero-padded cycling data with a dimension of ($3\times T$) and set its last layer as the output layer.

Additionally, considering cycling data are multivariate time series, we also study the effectiveness of models \cite{DBLP:conf/iclr/Wang0HWCX23,DBLP:conf/iclr/NieNSK23,DBLP:conf/nips/WuXWL21,DBLP:conf/aaai/ZengCZ023,DBLP:conf/iclr/LiuHZWWML24}, which are originally developed for other multivariate time series, for BLP. Specifically, we further include five baselines: DLinear \cite{DBLP:conf/aaai/ZengCZ023}, PatchTST \cite{DBLP:conf/iclr/NieNSK23}, Autoformer \cite{DBLP:conf/nips/WuXWL21}, iTransformer \cite{DBLP:conf/iclr/LiuHZWWML24} and MICN \cite{DBLP:conf/iclr/Wang0HWCX23}. The implementation details for the popular benchmark methods are described in Appendix~\ref{AppendixImplementation}. A dummy model that uses the mean of training labels as the prediction is also included as a baseline.

\begin{figure}[t]
	\centering
	\includegraphics[width=0.75\linewidth]{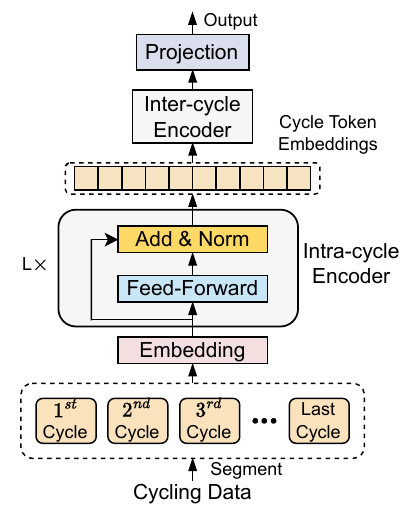}
	\caption{Overview of the CyclePatch framework.}
	\Description[Cyclepatch]{Overview of the CyclePatch framework.}
	\label{FigCyclePatch}
\end{figure}

\subsection{CyclePatch}
In addition to the popular baselines presented in Section~\ref{SecBaselines}, we propose CyclePatch, a simple yet effective plug-in technique, as shown in Figure~\ref{FigCyclePatch}. CyclePatch treats each cycle as a token:
\begin{gather}
	\left[\textbf{X}_1, \textbf{X}_2, \textbf{X}_3, \cdots, \textbf{X}_S \right] = {\rm Segment}(\textbf{X}_{1:S}), \\
	\hat{\textbf{X}}_i = \textbf{W} {\rm flatten}(\textbf{X}_i) + \textbf{b} ,\label{EqInitialEmbed}
\end{gather}
where $\textbf{W} \in \mathbb{R}^{D_1 \times 900}$, $\textbf{b} \in \mathbb{R}^{D_1}$, and ${\rm flatten}(\textbf{X}_i) \in \mathbb{R}^{900}$. In this framework, CyclePatch segments the cycling time series into basic units that have recurring patterns throughout degradation tests. As information of cycle tokens might differentiate due to changes in battery internal state and varied aging conditions, we leveraged an intra-cycle encoder to model information within each cycle. Specifically, we stack $L$ layers of feed-forward neural networks to deeply mine effective patterns within each cycle token. The computation at the $l^{th}$ layer of the intra-cycle encoder is given by:
\begin{gather}
	\hat{\textbf{z}}^l_i = \textbf{W}_2^l \sigma(\textbf{W}_1^l \textbf{z}^{l-1}_i + \textbf{b}_1^l) + \textbf{b}_2^l, \\
	\textbf{z}^l_i = LN\left(\hat{\textbf{z}}^l_i + \textbf{z}^{l-1}_i\right),
\end{gather}
where $\textbf{W}_1^l \in \mathbb{R}^{D_2 \times D_1}$, $\textbf{W}_2^l \in \mathbb{R}^{D_1 \times D_2}$, $\textbf{b}_1^l \in \mathbb{R}^{D_2}$, $\textbf{b}_2^l \in \mathbb{R}^{D_1}$, and $LN$ denotes layer normalization \cite{ba2016layernormalization}. $\textbf{z}^{l-1}_i$ is the input to the $l^{th}$ layer, and $\textbf{z}^0_i$ is initialized with $\hat{\textbf{X}}_i$. After process of all layers, the output $\textbf{z}^{L}_i$ from each cycle are concatenated:
\begin{gather}
	\textbf{H} = \text{Concat}(\textbf{z}^{L}_1, \textbf{z}^{L}_2, \cdots, \textbf{z}^{L}_S) \in \mathbb{R}^{S \times D_1}.
\end{gather}
An inter-cycle encoder is then applied to extract key patterns across cycle token embeddings:
\begin{gather}
	\textbf{v} = f(\textbf{H}), \\
	\hat{y} = {\rm Projection}(\textbf{v}) \label{EqLinearProjection},
\end{gather}
where $f(\cdot)$ represents the inter-cycle encoder. In this work, we explore the effectiveness of CyclePatch using various neural networks as the inter-cycle encoder, including MLP, Transformer encoder, LSTM, BiLSTM, GRU, and BiGRU. The vector $\textbf{v}$ captures both intra-cycle and inter-cycle information, and a linear projection produces the final prediction $\hat{y}$.

\section{EXPERIMENTS}
In this section, we report extensive experimental results wishing to answer the following research questions: \textbf{RQ1}: How do the benchmark methods perform on different domains? \textbf{RQ2}: How do the main components of CyclePatch framework affect the performance? \textbf{RQ3}: How adaptable are benchmark methods when applied across aging conditions in each domain? \textbf{RQ4}: How transferrable is the model pretrained on the Li-ion domain for other domains? 

\begin{table}[t]
	\caption{The data statistics of each part in BatteryLife}\label{TStatisticsEach}
	\resizebox{0.47\textwidth}{!}{
		\begin{tabular}{l|llll}
			\bottomrule
			& Li-ion & Zn-ion & Na-ion & \multicolumn{1}{l}{CALB} \\ \hline
			Battery Number        & 837  & 95    & 31    & 27                        \\
			Package Structure     & 5    & 3      & 1     & 1                         \\
			Chemical System       & 13   & 45     & 1     & 1                         \\
			Operation Temperature & 8    & 3      & 1    & 4                          \\
			Protocol   & 406  & 1      & 12      & 2                          \\
			Data Source           & Lab test  & Lab test  & Lab test  & Industrial test                       \\
			Battery Type          & Li-ion& Zn-ion       &Na-ion   & Li-ion                              \\ 
			\toprule
	\end{tabular}}
\end{table}

\begin{table*}[!t]
	\caption{Overall model performances. The top-three results are shown in shadow. The best results are shown in bold and the second results are underlined. The "-" represents out of memory. NA means the vanilla RNNs are not applicable for BLP because their speed for processing ultra-long time series is very slow and hence the model cannot converge even after training of twenty-four hours with eight GTX 4090 GPUs.} \label{TOverallPerformance}
	\resizebox{1.0\textwidth}{!}{
		\begin{tabular}{c|c|ccccccccc}
			\bottomrule
			\multicolumn{2}{c|}{Datasets} & \multicolumn{2}{c}{Li-ion} & \multicolumn{2}{c}{Zn-ion} & \multicolumn{2}{c}{Na-ion} & \multicolumn{2}{c}{CALB} & \\
			\hline
			\multicolumn{2}{c|}{Metrics} & MAPE & 15\%-Acc & MAPE & 15\%-Acc & MAPE & 15\%-Acc & MAPE & 15\%-Acc\\
			\hline
			\multicolumn{2}{c|}{Dummy} &$0.831_{\pm0.000}$ &$0.296_{\pm0.000}$ &$1.297_{\pm0.214}$ &$0.083_{\pm0.047}$ & $0.404_{\pm0.029}$ & $0.067_{\pm0.094}$ &$1.811_{\pm0.550}$ &$0.267_{\pm0.094}$ \\
			\hline
			\multirow{3}{*}{MLPs}
			& DLinear &$0.586_{\pm0.028}$ &$0.275_{\pm0.017}$ &$0.814_{\pm0.026}$ &$0.124_{\pm0.020}$ & $0.319_{\pm0.031}$ & $0.329_{\pm0.042}$ &$0.164_{\pm0.049}$ &$0.601_{\pm0.114}$ \\
			& MLP &$0.233_{\pm0.010}$ &$0.503_{\pm0.013}$ &$0.805_{\pm0.103}$ &$0.079_{\pm0.055}$ & $0.281_{\pm0.067}$ & $0.364_{\pm0.098}$ &$0.149_{\pm0.014}$ &$0.641_{\pm0.115}$ \\
			& CPMLP &\cellcolor{gray!20}{\textbf{\boldmath{$0.179_{\pm0.003}$}}} &\cellcolor{gray!20}\textbf{\boldmath{$0.620_{\pm0.004}$}} &\cellcolor{gray!20}\underline{$0.558_{\pm0.034}$} &\cellcolor{gray!20}{\textbf{\boldmath{{$0.297_{\pm0.084}$}}}} & \cellcolor{gray!20}$0.274_{\pm0.026}$ & $0.337_{\pm0.038}$ &\cellcolor{gray!20}\textbf{\boldmath{$0.140_{\pm0.009}$}} &\cellcolor{gray!20}\textbf{\boldmath{$0.704_{\pm0.053}$}} \\
			\hline
			\multirow{5}{*}{Transformers}
			& PatchTST &$0.288_{\pm0.042}$ &$0.430_{\pm0.053}$ &$0.716_{\pm0.024}$ &$0.133_{\pm0.001}$ & $0.396_{\pm0.094}$ & $0.258_{\pm0.070}$ &$0.347_{\pm0.045}$ &$0.511_{\pm0.139}$ \\
			& Autoformer &$0.437_{\pm0.093}$ &$0.287_{\pm0.067}$ &$0.987_{\pm0.243}$ &$0.106_{\pm0.039}$ & $0.372_{\pm0.047}$ & $0.177_{\pm0.128}$ &$0.761_{\pm0.061}$ &$0.329_{\pm0.121}$ \\
			& iTransformer &$0.209_{\pm0.015}$ &$0.516_{\pm0.028}$ &$0.690_{\pm0.110}$ &$0.188_{\pm0.037}$ & $0.321_{\pm0.087}$ & $0.249_{\pm0.178}$ &$0.164_{\pm0.020}$ &  $0.649_{\pm0.044}$ \\
			& Transformer &- &- &- &- &- &- &- &- \\
			& CPTransformer &\cellcolor{gray!20}\underline{$0.184_{\pm0.003}$} &$0.573_{\pm0.016}$ &\cellcolor{gray!20}{\textbf{\boldmath{$0.515_{\pm0.067}$}}} &$0.202_{\pm0.084}$ & \cellcolor{gray!20}\textbf{\boldmath{$0.255_{\pm0.036}$}} & \cellcolor{gray!20}\textbf{\boldmath{$0.406_{\pm0.084}$}} &\cellcolor{gray!20}$0.149_{\pm0.005}$ &$0.672_{\pm0.107}$ \\
			\hline
			\multirow{2}{*}{CNNs}
			& CNN &$0.337_{\pm0.068}$ &$0.371_{\pm0.050}$ &$0.928_{\pm0.093}$ &$0.115_{\pm0.029}$ & $0.307_{\pm0.047}$ & $0.273_{\pm0.027}$ &$0.278_{\pm0.011}$ &$0.582_{\pm0.032}$ & \\
			& MICN &$0.249_{\pm0.004}$ &$0.494_{\pm0.019}$ &\cellcolor{gray!20}$0.579_{\pm0.101}$ &\cellcolor{gray!20}$0.227_{\pm0.127}$ & $0.305_{\pm0.040}$ & $0.335_{\pm0.065}$ &$0.233_{\pm0.050}$ &$0.471_{\pm0.257}$ &\\
			\hline
			\multirow{6}{*}{RNNs}
			& CPGRU &\cellcolor{gray!20}$0.189_{\pm0.008}$ &\cellcolor{gray!20}\underline{$0.585_{\pm0.013}$} &$0.616_{\pm0.049}$ &\cellcolor{gray!20}\underline{$0.289_{\pm0.076}$} & $0.298_{\pm0.063}$ & $0.203_{\pm0.160}$ &\cellcolor{gray!20}\underline{$0.141_{\pm0.012}$} &\cellcolor{gray!20}$0.681_{\pm0.178}$ \\
			& CPBiGRU &$0.190_{\pm0.001}$ &{$0.566_{\pm0.034}$} &$0.774_{\pm0.202}$ &$0.193_{\pm0.156}$ & $0.282_{\pm0.055}$ & \cellcolor{gray!20}$0.395_{\pm0.008}$ &$0.160_{\pm0.015}$ &\cellcolor{gray!20}\underline{$0.686_{\pm0.063}$} \\
			& CPLSTM &$0.196_{\pm0.006}$ &\cellcolor{gray!20}$0.585_{\pm0.020}$ &$0.932_{\pm0.227}$ &$0.085_{\pm0.028}$ & \cellcolor{gray!20}\underline{$0.272_{\pm0.051}$} & $0.386_{\pm0.009}$ &$0.156_{\pm0.073}$ &$0.613_{\pm0.153}$ \\
			& CPBiLSTM &$0.191_{\pm0.007}$ &$0.421_{\pm0.255}$ &$0.645_{\pm0.049}$ &$0.150_{\pm0.104}$ & $0.299_{\pm0.043}$ & \cellcolor{gray!20}\underline{$0.399_{\pm0.001}$} &$0.173_{\pm0.075}$ &$0.663_{\pm0.247}$ \\
			& GRU\&BiGRU & NA & NA & NA &NA & NA & NA &NA &NA \\
			& LSTM\&BiLSTM & NA & NA & NA &NA & NA & NA &NA &NA \\
			\toprule
	\end{tabular}}
\end{table*}

\begin{figure*}[!ht]
	\centering
	\includegraphics[width=0.9\linewidth]{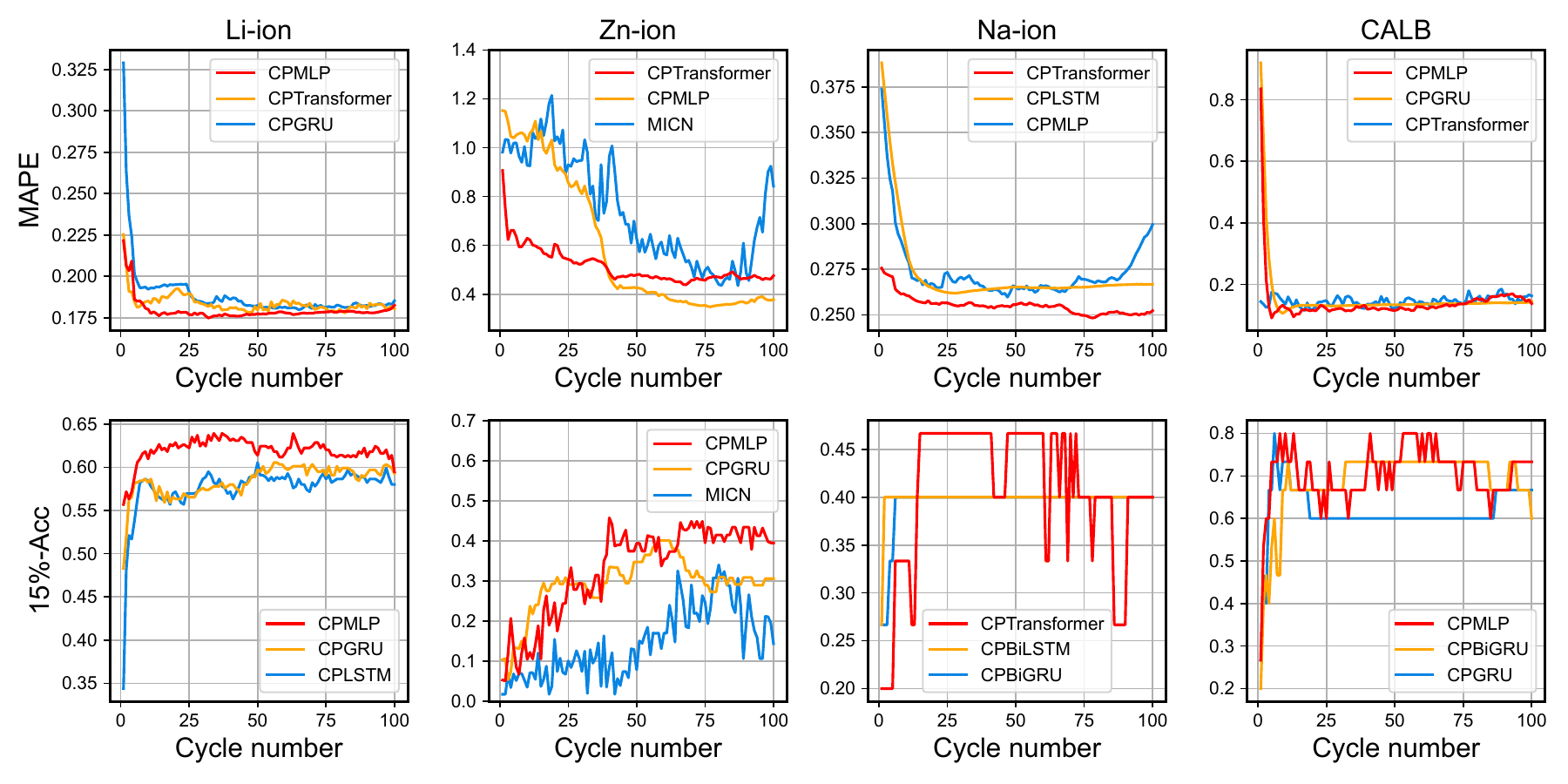}
	\caption{The performance of the top-three models with an increasing number of usable cycles.}
	\Description[error curve]{The performance of the top-three models with an increasing number of usable cycles.}
	\label{FigErrorCurve}
\end{figure*}
\subsection{Experimental Setup}\label{SecExperimentalDetails}
The raw cycling data vary in the number of time steps across cycles and batteries. To ensure consistency, for each cycle, we use linear interpolation to resample both charging and discharging data to 150 points, each containing current, voltage, and capacity records. In this vein, each cycle comprises 300 data points. This results in up to 30,000 data points in samples when the first 100 cycles are used. Prior to model input, the resampled cycling data are normalized as detailed in Appendix~\ref{AppendixImplementation}.

For most batteries in BatteryLife, we use the measured end-of-life as the life label. However, a few batteries remain above 80\% SOH at the end of cycling. While some prior works \cite{20223412623287,20243216836174} use the last recorded cycle number as the life label, this can introduce large label noise. To mitigate this issue, we exclude batteries with a final SOH greater than $\lambda+2.5\%$. For those with a SOH in ($\lambda$, $\lambda+2.5\%$], we apply linear extrapolation to estimate the life label at $\lambda$ SOH. In this work, $\lambda = 80\%$ for all datasets except CALB, where $\lambda = 90\%$ is used, as most of the collected CALB batteries do not reach 80\% SOH. Additionally, for CALB batteries, we use $Q_1$ as $Q_0$ in Equation~\ref{EqSOH} to calculate SOH in accordance with CALB regulations.

\begin{table*}[t]
	\centering
	\caption{Results of ablation studies. The "-" represents out of memory.}
	\label{TAblation}
	\resizebox{1.0\textwidth}{!}{
		\begin{tabular}{c|l|cccccccc}
			\bottomrule
			\hline
			\multirow{2}{*}{Models}        & \multicolumn{1}{c|}{\multirow{2}{*}{Ablations}} & \multicolumn{2}{c}{Li-ion} & \multicolumn{2}{c}{Zn-ion} & \multicolumn{2}{c}{Na-ion} & \multicolumn{2}{c}{CALB}  \\ \cline{3-10} 
			& \multicolumn{1}{c|}{}                           & MAPE         & 15\%-ACC    & MAPE         & 15\%-ACC    & MAPE         & 15\%-ACC    & MAPE        & 15\%-ACC    \\ \hline
			\multirow{4}{*}{CPMLP}         & Full                                            & \boldmath{$0.179_{\pm0.003}$}  & \boldmath{$0.620_{\pm0.004}$} & \boldmath{$0.558_{\pm0.034}$}  & \boldmath{$0.297_{\pm0.084}$} & \boldmath{$0.274_{\pm0.026}$}  & $0.337_{\pm0.038}$ & \boldmath{$0.140_{\pm0.009}$} & \boldmath{$0.704_{\pm0.053}$} \\
			& w/o Intra                                       & $0.202_{\pm0.005}$  & $0.544_{\pm0.007}$ & $0.585_{\pm0.147}$  & $0.277_{\pm0.102}$ & $0.345_{\pm0.091}$  & $0.208_{\pm0.154}$ & $0.157_{\pm0.015}$ & $0.683_{\pm0.027}$ \\
			& w/o Inter                                       & $0.199_{\pm0.016}$  & $0.560_{\pm0.015}$ & $0.761_{\pm0.062}$  & $0.187_{\pm0.012}$ & $0.357_{\pm0.091}$  & $0.194_{\pm0.156}$ & $0.203_{\pm0.098}$ & $0.565_{\pm0.170}$ \\
			& MLP                                             & $0.233_{\pm0.010}$  & $0.503_{\pm0.013}$ & $0.805_{\pm0.103}$  & $0.079_{\pm0.055}$ & $0.281_{\pm0.067}$  & \boldmath{$0.364_{\pm0.098}$} & $0.149_{\pm0.014}$ & $0.641_{\pm0.115}$ \\ \hline 
			\multirow{4}{*}{CPTransformer} & Full                                            & \boldmath{$0.184_{\pm0.003}$}  & \boldmath{$0.573_{\pm0.016}$} & \boldmath{$0.515_{\pm0.067}$}  & \boldmath{$0.202_{\pm0.084}$} & \boldmath{$0.255_{\pm0.036}$}  & \boldmath{$0.406_{\pm0.084}$} & \boldmath{$0.149_{\pm0.005}$} & \boldmath{$0.672_{\pm0.107}$} \\
			& w/o Intra                                       & $0.197_{\pm0.005}$  & $0.542_{\pm0.018}$ & $0.530_{\pm0.030}$  & $0.152_{\pm0.109}$ & $0.337_{\pm0.084}$  & $0.249_{\pm0.181}$ & $0.214_{\pm0.036}$ & $0.483_{\pm0.182}$ \\
			& w/o Inter                                       & $0.188_{\pm0.003}$  & $0.568_{\pm0.011}$ & $0.655_{\pm0.028}$  & $0.114_{\pm0.014}$ & $0.287_{\pm0.070}$  & $0.279_{\pm0.181}$ & $0.205_{\pm0.028}$ & $0.563_{\pm0.022}$ \\
			& Transformer                                     & -            & -           & -            & -           & -            & -           & -           & -           \\ \hline
			\toprule
	\end{tabular}}
\end{table*}
We randomly split BatteryLife into training, validation, and testing sets with a 6:2:2 ratio. All models are trained to minimize mean squared error using the Adam optimizer \cite{adam}. Note that models utilizing CyclePatch are prefixed with "CP" in their names. For instance, CPTransformer employs CyclePatch and uses a Transformer encoder as the inter-cycle encoder. Transformer refers to the Transformer encoder in this work. We conduct more than 3,000 hyperparameter searches in total (details in Appendix~\ref{AppendixHyperpara}), and select the best-performing hyperparameters based on the validation set. To reduce the impact of experimental fluctuations, we run each experiment three times and report the mean$\pm$standard deviation on testing sets for all evaluations.

\subsection{Evaluation Metrics}
We employ two metrics to evaluate model performance: mean absolute percentage error (MAPE) and $\alpha$-accuracy \cite{20211510188187}. MAPE is widely adopted as the main evaluation metric for BLP \cite{20191406719359,20243216836174,20223412623287,FEI2023106903,TANG2022111530}, and computed as:
\begin{equation}
	{\rm MAPE}=\frac{1}{N}\sum_{i}^{N}\frac{\left|y_i-\hat{y}_i\right|}{y_i},
\end{equation}
where $y_i$ and $\hat{y}_i$ are the ground truth and predicted battery life of the $i^{th}$ sample, respectively. $N$ is the number of samples in the testing set. As for $\alpha$-accuracy, it has been employed to assess model performance from an engineering point of view in the battery filed \cite{20211510188187,TAN2024103725}, and is written by:
\begin{equation}
	\alpha{\rm -accuracy}=\frac{1}{N}\sum_{i=1}^{N}\textbf{1}_{\left|y_i-\hat{y}_i\right|\leq\alpha y_i}(\hat{y}_i),
\end{equation}
where $\textbf{1}_{\left|y_i-\hat{y}_i\right|\leq\alpha y_i}(\hat{y}_i)$ is an indicator function that produces 1 if $\left|y_i-\hat{y}_i\right|\leq\alpha y_i$ holds. In this work, we consider predictions usable when the relative error is no larger than 15\%, and thus use $\alpha=15\%$ to reflect the percentage of usable prediction in the testing data. The $15\%{\rm-accuracy}$ is abbreviated as $15\%{\rm-Acc}$ in this paper.

\subsection{Model Performance on Different Domains (RQ1)}
In this subsection, we compare the performance of benchmark methods on different domains. The benchmark methods are categorized into four categories (MLPs, Transformers, CNNs, and RNNs) and their overall model performance is reported in Table~\ref{TOverallPerformance}. It is observed that MLPs and Transformers generally outperform other categories. This phenomenon matches the modern landscape in other time series tasks \cite{DBLP:conf/aaai/ZengCZ023,DBLP:conf/iclr/NieNSK23,DBLP:conf/nips/WuXWL21,DBLP:conf/iclr/LiuHZWWML24, 20241515882168}. Remarkably, CyclePatch methods achieve the best performance across all domains, underscoring the effectiveness of the proposed plug-in technique. We also observe that vanilla Transformer and RNNs are not suitable for BLP due to memory limitations or slow training speeds when handling sequences with 30,000 points. In contrast, CyclePatch enables these models to become viable for this task, improving memory efficiency and training speed. Additionally, the best MAPE are 0.179, 0.515, 0.255, and 0.141 for the Li-ion, Zn-ion, Na-ion and CALB datasets, respectively. The poor performance on the Zn-ion dataset can be attributed to the challenge of learning diverse chemical systems with limited data available for each system.

With respect to approaches popular for other time series data \cite{DBLP:conf/iclr/Wang0HWCX23,DBLP:conf/iclr/NieNSK23,DBLP:conf/nips/WuXWL21,DBLP:conf/aaai/ZengCZ023,DBLP:conf/iclr/LiuHZWWML24}, they generally underperform for BLP. The possible reasons are two-fold. On the one hand, the explicit architectural designs used for decomposing trends and seasonal patterns are not suitable for BLP. This is mainly supported by the evidence that models with trend and seasonal decomposition (DLinear, Autoformer, and MICN) perform poorly on most datasets. This is further corroborated by the results that MLP generally outperforms DLinear with clear gaps, whereas MLP removes the decomposition design from DLinear. The failure of trend and seasonal decomposition is reasonable, as the life-related information underlies the fine-grained interactions among variables (voltage, current, and capacity) rather than trend and seasons, as illustrated in Figure~\ref{demoFigure}. Moreover, since cycling data have regular seasons (cycles), learning these seasons becomes redundant.

On the other hand, certain techniques are suboptimal for battery cycling data. Particularly, as shown in Table~\ref{TOverallPerformance} and Table~\ref{TAblation}, it is clear that CPTransformer without the intra-cycle encoder still consistently outperforms PatchTST while CPTransformer without the intra-cycle encoder is a special case for PatchTST without RevIN \cite{20231213775332}. This indicates that RevIN, which is commonly adopted in time series models \cite{DBLP:conf/iclr/NieNSK23,DBLP:conf/nips/WangWDQZLQWL24,DBLP:conf/iclr/WangWSHLMZ024,DBLP:conf/iclr/WuHLZ0L23}, destroys the inherent interaction among variables. As for iTransformer \cite{DBLP:conf/iclr/LiuHZWWML24}, it embeds the time series of a variate into a variate token and then models the interaction among variate tokens, resulting in loss of fine-grained interactions at different temporal regions. Additionally, while patching is commonly used in time series tasks, existing approaches typically employ a simple linear layer to generate token embeddings. In contrast, CyclePatch operates on cycle tokens and applies an intra-cycle encoder to capture intra-cycle interactions before applying inter-cycle encoders, preserving the intrinsic characteristics of the cycling data. These findings indicate that techniques successful in other time series fields cannot be naively applied to BLP.

We also investigate the performance of the top-three models as the number of usable cycles increases across different domains, with the results shown in Figure~\ref{FigErrorCurve}. The model performance improves with the increase in the number of usable cycles initially, then plateaus. This indicates that while more cycles provide additional information, the information gain diminishes as the number of cycles exceeds a certain threshold. Moreover, we note that the best benchmark methods for Zn-ion and CALB datasets may not maintain the best performance with certain cycle numbers. These findings suggest that developing models capable of effectively handling cycling data with various cycle numbers could be a promising avenue for future research.

\begin{table}[!t]
	\caption{Results of model performance on seen and unseen aging conditions. The best results on seen and unseen aging conditions are highlighted.} \label{TSeenUnseen}
	\centering
	\resizebox{0.47\textwidth}{!}{
		\begin{tabular}{c|c|c|cc}
			\bottomrule
			\multicolumn{1}{c|}{Datasets} &\multicolumn{1}{c|}{Models} &\multicolumn{1}{c|}{Metrics}  & Seen & Unseen \\ \hline
			\multirow{6}{*}{Li-ion}
			&\multirow{2}{*}{CPGRU} 
			& MAPE & $0.175_{\pm0.011}$ & $0.203_{\pm0.005}$\\
			~ & ~ & 15\%-Acc & $0.626_{\pm0.019}$ & $0.541_{\pm0.010}$\\ \cline{2-5}
			&\multirow{2}{*}{CPTransformer} 
			& MAPE & \boldmath{$0.157_{\pm0.003}$} & $0.211_{\pm0.003}$\\ 
			~ & ~ & 15\%-Acc & $0.608_{\pm0.034}$ & $0.535_{\pm0.013}$\\  \cline{2-5}
			&\multirow{2}{*}{CPMLP} 
			& MAPE & $0.158_{\pm0.003}$ & \boldmath{$0.203_{\pm0.004}$} \\ 
			~ & ~ & 15\%-Acc & \boldmath{$0.635_{\pm0.013}$} & \boldmath{$0.603_{\pm0.008}$}\\ 
			\hline
			\hline
			\multirow{6}{*}{Zn-ion} 
			&\multirow{2}{*}{MICN} 
			& MAPE & $0.721_{\pm0.069}$ & $0.901_{\pm0.330}$\\ 
			~ & ~ & 15\%-Acc & $0.148_{\pm0.060}$ & $0.111_{\pm0.023}$\\ \cline{2-5}
			&\multirow{2}{*}{CPMLP} 
			& MAPE & $0.591_{\pm0.120}$ & $0.600_{\pm0.074}$\\ 
			~ & ~ & 15\%-Acc & \boldmath{$0.327_{\pm0.064}$} & \boldmath{$0.388_{\pm0.151}$}\\ \cline{2-5}
			&\multirow{2}{*}{CPTransformer} 
			& MAPE & \boldmath{$0.545_{\pm0.150}$} & \boldmath{$0.485_{\pm0.168}$}\\
			~ & ~ & 15\%-Acc & $0.199_{\pm0.136}$ & $0.191_{\pm0.030}$\\ 
			\hline
			\hline
			\multirow{6}{*}{Na-ion} 
			&\multirow{2}{*}{CPMLP} 
			& MAPE & $0.192_{\pm0.042}$ & $0.397_{\pm0.025}$ \\ 
			~ & ~ & 15\%-Acc & $0.561_{\pm0.064}$ & $0.000_{\pm0.000}$  \\ \cline{2-5}
			&\multirow{2}{*}{CPLSTM}
			& MAPE & $0.210_{\pm0.069}$ & $0.432_{\pm0.004}$  \\ 
			~ & ~ & 15\%-Acc & \boldmath{$0.664_{\pm0.002}$} & $0.000_{\pm0.000}$ \\ \cline{2-5}
			&\multirow{2}{*}{CPTransformer} 
			& MAPE & \boldmath{$0.186_{\pm0.049}$} & \boldmath{$0.359_{\pm0.020}$} \\ 
			~ & ~ & 15\%-Acc & $0.540_{\pm0.177}$ & \boldmath{$0.205_{\pm0.199}$} \\ 
			\hline
			\hline
			\multirow{6}{*}{CALB} 
			&\multirow{2}{*}{CPTransformer} 
			& MAPE & $0.138_{\pm0.044}$ & $0.141_{\pm0.075}$\\
			~ & ~ & 15\%-Acc & $0.696_{\pm0.095}$ & $0.660_{\pm0.226}$\\ \cline{2-5}
			&\multirow{2}{*}{CPGRU} 
			& MAPE & $0.160_{\pm0.009}$ & \boldmath{$0.127_{\pm0.047}$} \\ 
			~ & ~ & 15\%-Acc & $0.595_{\pm0.181}$ & \boldmath{$0.692_{\pm0.219}$}\\ \cline{2-5}
			&\multirow{2}{*}{CPMLP} 
			& MAPE & \boldmath{$0.112_{\pm0.026}$} & $0.178_{\pm0.059}$ \\ 
			~ & ~ & 15\%-Acc & \boldmath{$0.796_{\pm0.048}$} & $0.602_{\pm0.254}$\\ 
			\toprule
		\end{tabular}
	}
\end{table}
\subsection{Ablation Studies (RQ2)}
We conduct ablation studies of the proposed CyclePatch to study the effectiveness of its key components, and present the results in Table~\ref{TAblation}. Specifically, we study CPMLP and CPTransformer considering their leading performance across datasets. "w/o Intra" refers to the removal of the intra-cycle encoder; "w/o Inter" indicates the absence of the inter-cycle encoder; MLP and Transformer severs as naive baselines without patching. From Table~\ref{TAblation}, it is evident that all components in CyclePatch significantly contribute to its performance. Concretely, the incorporation of the intra-cycle encoder enhances the model's ability to model interactions among variables within cycles, leading to performance gains across all datasets. Removing the inter-cycle encoder reduces performance, highlighting the importance of modeling inter-cycle interactions. Moreover, it is observed that CPMLP without either intra-encoder or inter-encoder performs worse than MLP on Na-ion and CALB datasets, whereas MLP naively models the intra-cycle and inter-cycle information. This indicates that modeling intra-cycle and inter-cycle information are both critical. These results collectively validate the effectiveness of the CyclePatch design.

\subsection{Model Performance on Seen and Unseen Aging Conditions (RQ3)}
We further investigate the generalizability of the top-three models by evaluating their performance on aging conditions that are seen and unseen in the training set, and present results in Table~\ref{TSeenUnseen}. This is crucial because a broad range of BLP applications demands models that remain effective across aging conditions. The results indicate that models generally perform worse on unseen aging conditions compared to those seen ones. This discrepancy arises from domain shifts, which result from different data distributions across aging conditions, as discussed in Section~\ref{SecDegradationTest}. Additionally, we note that models that excel in either seen or unseen aging conditions may not be the best models
concerning overall performance. This indicates that batteryLife requires models capable of handling independent identical distribution and out of distribution simultaneously.

\begin{table}[!t]
	\caption{Results of cross-domain transferability.}
	\label{TCrossDomain}
	\centering
	\resizebox{0.47\textwidth}{!}{
		\begin{tabular}{c|c|cccc}
			\bottomrule
			\multicolumn{1}{l|}{Methods} & Metrics  & Baseline   & Frozen & FT & DA \\ \hline
			\multirow{2}{*}{Zn-ion}      & MAPE     & \boldmath{$0.515_{\pm0.067}$} & $3.153_{\pm0.187}$       & $0.562_{\pm0.125}$    & $0.752_{\pm0.102}$   \\
			& 15\%-ACC & \boldmath{$0.297_{\pm0.084}$}  & $0.046_{\pm0.028}$        & $0.301_{\pm0.093}$    & $0.070_{\pm0.033}$    \\ \hline
			\multirow{2}{*}{Na-ion}      & MAPE     & \boldmath{$0.255_{\pm0.036}$} & $4.115_{\pm0.138}$ & $0.286_{\pm0.034}$ & $0.641_{\pm0.121}$   \\
			& 15\%-ACC & \boldmath{$0.406_{\pm0.084}$} & $0.000_{\pm0.000}$ &$0.341_{\pm0.072}$ & $0.263_{\pm0.081}$   \\ \hline
			\multirow{2}{*}{CALB}        & MAPE     & \boldmath{$0.140_{\pm0.009}$} & $1.536_{\pm0.147}$       &  $0.146_{\pm0.051}$  & $0.256_{\pm0.070}$   \\
			& 15\%-ACC &   $0.704_{\pm0.053}$          & $0.219_{\pm0.109}$       &  \boldmath{$0.767_{\pm0.088}$}   & $0.404_{\pm0.052}$  \\ \toprule
	\end{tabular}}
\end{table}

\subsection{Cross-domain Transferability (RQ4)}
In real-world situations, batteries of new types are often fewer than batteries of old types so that it is important to develop a model that has cross-domain transferability against the domain shifts caused by battery types. In this work, we evaluate the cross-domain transferability of models by investigating if a model pretrained on Li-ion dataset can be transferred to other domains, and report the results in Table~\ref{TCrossDomain}. Specifically, we study the transferability of CPMLP which is the best model in Li-ion domain. Three transferability evaluations are considered: (1) Frozen: Directly applying the pretrained CPMLP on target domains. (2) Fine-tuning (FT): Fine-tune all parameters of the pretrained CPMLP. (3) Domain adaptation (DA): we implement a domain adaptation technique by minimizing maximum mean discrepancy \cite{HAN2022230823} between the target-domain and source-domain embeddings produced by the last hidden layer. By expectation, with transfer learning techniques, an effective transferrable model should outperform the best baseline that is trained using only target-domain training data.

Table~\ref{TCrossDomain} shows that frozen pretrained models perform poorly on target domains due to significant domain shifts. While fine-tuning and domain adaptation improve performance over frozen models, they underperform the baseline in most cases. This suggests that extracting relevant information from diverse lab-test data (with hundreds of aging conditions) to other battery types remains challenging. The cross-domain evaluation results indicate that BatteryLife still has substantial room for improvements in cross-domain transferability.

\section{CONCLUSION AND FUTURE WORK}
In this paper, we propose BatteryLife, which offers the largest and most diverse battery life dataset created by integrating 16 datasets. The unprecedented dataset size and diversity enable advances in BatteryLife to produce actionable insights for the broad battery and data mining communities. Moreover, BatteryLife provides a comprehensive benchmark, including baselines popular in this and other time series fields. Extensive experiments demonstrate that BatteryLife contains research opportunities for developing tailored time series models for BLP. In future studies, we also focus on the research areas discovered in this work and plan to incorporate more datasets into BatteryLife.

\section{ACKNOWLEDGMENTS}
The authors acknowledge the financial support of the National
Key R\&D Program of China (No. 2023YFB2503600). This
work is also supported by research grants from the National
Natural Science Foundation of China (No. 52207230, No. 92372109 and No. 62206067), the Guangzhou-HKUST(GZ) Joint Funding
Program (No. 2023A03J0003, No. 2023A03J0103 and No. 2023A03J0673), and
the Guangzhou Municipal Science and Technology Project
(No. 2024A04J4216).

\bibliographystyle{ACM-Reference-Format}
\bibliography{sample-sigconf}

\appendix

\section{FURTHER DETAILS ABOUT THE DATASETS}\label{AppendixA}
\subsection{Data Preprocessing}\label{Preprocessing}
In this section, we introduce the preprocessing details for all datasets. We obtained HNEI, MICH, MICH\_EXP, SNL, and UL\_PUR datasets from BatteryArchive. Other datasets were sourced from the links shared in the original papers. For all datasets, the present study filtered out the batteries whose SOH is not degraded to $\lambda+2.5\%$ ($\lambda=90\%$ for the CALB dataset and $\lambda=80\%$ for other datasets), and then conducted the data statistics. Note that (1) BatteryArchive is an active website, and the data statistics were conducted on the available data until September 2024 in this paper. (2) Stanford~\cite{20244517308214} were released multiple times and the data used in this work is from the first release. As for BatteryLife, we further excluded the batteries whose life labels are no larger than 100. Additionally, we found that some batteries in the UL\_PUR dataset contain sudden significant SOH drops without recovery. We conjecture that these drops are caused by equipment faults and thus removed these batteries. The data statistics were conducted on the remaining batteries. The statistics of BatteryLife were conducted on the remaining batteries. 

Furthermore, some outliers might exist in the cycling data. For the 7 datasets standardized by BatteryML \cite{20243216836174}, we detected abnormal cycles by running a median filtering method on discharge capacity degradation trajectory following BatteryML \cite{20243216836174} and removed the detected outliers. As for other datasets, we first removed the rest performance tests (RPT) and formation cycles. The RPT cycles are removed because they comply with charge/discharge protocols that are different from the cycling protocols, and the RPT cycles are also generally detected as outliers by BatteryML \cite{20243216836174}. Likewise, formation cycles are removed. Moreover, some cycles of sudden rise or drop in discharge capacities are manually removed. After data cleaning, each battery was stored in a data format consistent with BatteryML \cite{20243216836174}.

\subsection{Degradation Test Details}\label{PrivateDetails}
The data for the zinc-ion batteries were collected using custom-made coin batteries fabricated in the laboratory. The coin batteries utilized stainless steel CR2032, CR2025 and CR2016 battery cases sourced from Guangdong Canrd New Energy Technology Co., Ltd., paired with stainless steel spring washers with a height of 0.6 mm and stainless  steel gaskets of 0.5 mm thickness, all sealed under a pressure of 75 kg/cm². The cathode material of the zinc battery was commercial manganese dioxide cathodes purchased from Guangdong Canrd New Energy Technology Co., Ltd. The anode material consisted of 99.99\% pure zinc foil produced by a local manufacturer. The separator used was Glass Fiber-D. The electrolyte employed was an aqueous solution of 2 mol/L zinc sulfate and 0.1 mol/L manganese sulfate, with various additives of different types and concentrations. The battery tests were conducted at a current density of 400 mA/m² and evaluated at temperatures of 25°C, 30°C, and 40°C.

The high-rate sodium-ion 18650 cylindrical batteries utilized in this study were sourced from Zhuhai Punashidai New Energy Co., Ltd., and were employed without any additional processing. The specific composition of the electrode materials and electrolyte remains proprietary information of the manufacturer. The nominal capacity of these batteries was determined through charge and discharge cycles at a 1C rate. All cycling tests were conducted within a voltage range of 2.0 V to 4.0 V for both charging and discharging processes. To assess the long-term performance, cycling tests were carried out at 12 distinct current rates, ranging from 2C to 6C, under one operation temperature: 25°C. 

The data for the lithium-ion batteries in the CALB dataset were provided by CALB Group Co., Ltd.. Prior to testing, all batteries were placed until thermal equilibrium. Two different protocols were then applied. a) For any cycle of batteries tested at 0°C, the batteries were discharged at a constant current of 1C until the voltage of batteries reached 2.2 V. The batteries were then rested for 10 minutes. After that, the batteries were charged at a constant current of 1 C to 4.35 V and then charged at a constant voltage of 4.35 V until the current was lower than 0.05C. The batteries were then rested for another 10 minutes. It should be noted that this constitutes an accelerated degradation test, which exceeds the design specifications of these batteries. b) For any cycle of batteries tested at 25°C, 35°C and 45°C, the batteries undergone 9 different charging stages. In the first stage, the batteries were charged at a constant current of 1.2C to 3.83 V and rested for 10 minutes after the first charging stage. In the second stage, the batteries were charged at a constant current of 1C to 3.89 V. In the third stage, the batteries were charged at a constant current of 0.85C to 3.95 V. In the fourth stage, the batteries were charged at a constant current of 0.75C to 4.08 V. In the fifth stage, the batteries were charged at a constant current of 0.6C to 4.11 V. In the sixth stage, the batteries were charged at a constant current of 0.5C to 4.19 V. In the seventh stage, the batteries were charged at a constant current of 0.33C to 4.24 V. In the eighth stage, the batteries were charged at a constant current of 0.15C to 4.35 V.  In the ninth stage, the batteries were charged at a constant voltage of 4.35 V until the current was lower than 0.05C. The batteries were then discharged at a constant current of 1C until reaching 2.75 V. Then, the batteries were rested for another 10 minutes.

\subsection{Overview of The Integrated Datasets}\label{OverviewIntegration}
We briefly introduce the integrated datasets below:
\begin{itemize}[left=0pt]
	\item CALCE \cite{20132816492471, HE201110314}: This dataset consists of 13 prismatic lithium-ion batteries, with $\rm LiCoO_{2}$ as the positive electrode and graphite as the negative electrode, while the electrolyte is unknown. The nominal capacity is 1.1 Ah or 1.35 Ah. The working temperature of all batteries is 25 degree Celsius. 2 different protocols are included totally in this dataset.
	
	\item HNEI \cite{20182105222726}: In this dataset, there are 14 18650-format lithium-ion batteries (LG Chem, ICR18650 C2) whose cathode is a mixture of $\rm LiCoO_{2}$ and $\rm LiNi_{0.4}Co_{0.4}Mn_{0.2}O_{2}$, and the anode is graphite. The electrolyte composition is not disclosed, and the nominal capacity is 2.8 Ah for all batteries. There are 2 different protocols with a working temperature of 25 degree Celsius.
	
	\item MATR \cite{attia2020closed, 20191406719359}: This dataset consists of 169 commercial Li-ion batteries in a format of 18650 cylindrical batteries whose positive electrode is lithium iron phosphate ($\rm LiFePO_{4}$), the negative electrode is graphite, and the electrolyte is unknown. The nominal capacity of all batteries is 1.1 Ah. All batteries were cycled under a working temperature of 30 degree Celsius with varied fast-charging protocols and identical discharge protocols. Totally, 81 different fast-charging protocols are included in this dataset.
	
	\item UL\_PUR \cite{juarez2020degradation}: This dataset covers 10 batteries in a format of 18650 cylindrical batteries. The cathode is $\rm LiNi_{0.8}Co_{0.15}Al_{0.05}O_{2}$. The anode is graphite. The electrolyte formula is unknown. The nominal capacity is 3.4 Ah. The working  temperature is 23 degree Celsius. All batteries underwent the same cycling protocol.
	
	\item SNL \cite{20203809192263}: This dataset consists of 50 batteries in the format of 18650 cylindrical batteries. The positive electrode covers $\rm LiFePO_{4}$, $\rm LiNi_{0.81}Co_{0.14}Al_{0.05}O_2$, and $\rm LiNi_{0.84}Mn_{0.06}Co_{0.1}O_2$, while the negative electrode is graphite. The electrolyte formula is unknown. There are 3 nominal capacities consisting of 1.1 Ah, 3.2 Ah, and 3.0 Ah. There are 23 different protocols in this dataset with 3 different working temperatures (15°C, 25°C and 35°C). 
	
	\item RWTH  \cite{LI2021230024}: This dataset has 48 lithium-ion batteries in the format of 18650 cylindrical batteries. The positive electrode is lithium nickel manganese cobalt oxide (NMC), and the negative electrode is carbon. The electrolyte is unknown. The nominal capacity is 3 Ah. All batteries are cycled with the same protocols with a working temperature of 25 degree Celsius.
	
	\item MICH\_EXP \cite{mohtat2021reversible}: This dataset consists of 18 lithium-ion batteries in the format of pouch batteries. The positive electrode is $\rm LiNi_{1/3}Co_{1/3}Mn_{1/3}O_2$ (NCM111), and the negative electrode is $\rm graphite$. The electrolyte consists of 1M $\rm LiPF_{6}$ salt in solvents of ethylene carbonate (EC) and ethyl methyl carbonate (EMC) with a ratio of 3:7. The nominal capacity is 5.0 Ah. There are 6 different protocols with 3 different working temperatures in total (-5°C, 25°C, and 45°C).
	
	\item MICH \cite{20214611166935}: This dataset consists of 40 lithium-ion batteries in the format of pouch batteries. The positive electrode is $\rm NCM111$ and the negative electrode is $\rm graphite$. The electrolyte consists of 1.0 M $\rm LiPF_{6}$ salt in solvents of EC and EMC  with a ratio of 3:7 together with 2 wt\% additive of vinylene carbonate. The nominal capacity is 2.36 Ah. There are 2 different protocols with 2 different working temperatures (25°C and 45°C).
	
	\item HUST \cite{20223412623287}: This dataset consists of 77 lithium-ion batteries in the format of 18650 cylindrical batteries. The positive electrode is lithium-iron-phosphate, and the negative electrode is graphite. The electrolyte for this dataset is unknown. The nominal capacity is 1.1 Ah. There are 77 different multi-stage discharge protocols but an identical fast-charging protocol with 30 degree Celsius as the working temperature.
	
	\item Tongji \cite{zhu2022data}: This dataset contains 108 commercial Li-ion batteries in a format of 18650 cylindrical batteries, covering 3 chemical systems consisting of 3 different positive electrodes, 2 different negative electrodes, and an identical electrolyte. This dataset includes 2 nominal capacities (3.5 Ah and 2.5 Ah), 3 operating temperatures, and 6 charge/discharge protocols.
	
	\item Stanford \cite{20244517308214}: This dataset covers 41 batteries in the format of pouch batteries whose positive electrode is $\rm LiNi_{0.5}Mn_{0.3}Co_{0.2}O_{2}$, negative electrode is artificial graphite. The electrolyte is a mixture of $\rm LiPF_6$, ethylene carbonate, ethyl methyl carbonate,  dimethyl carbonate, and vinylene carbonate. The nominal capacity of these batteries is 0.24 Ah. All batteries underwent different formation protocols but were cycled with an identical charge/discharge protocol under a working temperature of 30 degree Celsius.
	
	\item XJTU \cite{wang2024physics}: There are 23 batteries in this dataset in the format of 18650 cylindrical batteries. For all batteries, the positive electrode is $\rm LiNi_{0.5}Co_{0.2}Mn_{0.3}O_{2}$ and the negative electrode is graphite. The electrolyte is unknown. The nominal capacity is 2 Ah. This dataset includes 2 different operation conditions with a working temperature of 20 degree Celsius.
	
	\item ISU\_ILCC \cite{LI2024101891}: This dataset comprises 240 502030-size Li-polymer batteries. The positive electrode is lithium nickel manganese cobalt oxide, and the negative electrode is graphite. The electrolyte remains unknown. And the nominal capacity is 0.25 Ah. There are 202 different charge/discharge protocols in this dataset, with 30 degree Celsius as the working temperature.
	
	\item Zn-ion: This dataset consists of 100 zinc-ion batteries in the format of coin batteries. The positive electrode is $\rm MnO_{2}$ while the negative electrode is zinc metal. The electrolyte is unknown. The discharge capacity of the $10^{th}$ cycle was taken as the nominal capacity for each battery. All batteries underwent the same charge/discharge protocol at 3 different working temperatures.
	
	\item Na-ion: This dataset consists of 31 sodium-ion batteries in the format of 18650 cylindrical batteries. The positive electrode and the negative electrode are not disclosed. The electrolyte remains unknown. And the nominal capacity is 1.0 Ah. There are 12 different charge/discharge protocols in this dataset at 25 degree Celsius.
	
	\item CALB: This dataset consists of 27 lithium-ion batteries in the format of prismatic batteries. The positive electrode is NMC, and the negative electrode is graphite. The electrolyte is not disclosed. The nominal capacity is 58 Ah for all batteries. There are 2 different protocols with 4 different working temperatures (0°C, 25°C, 35°C and 45°C).
\end{itemize}

\section{FURTHER EXPERIMENTAL DETAILS}

\subsection{Further Details on Hyperparameters}\label{AppendixHyperpara}
For each model, we tried more than 20 different sets of hyperparameters and selected the hyperparameters that led to the lowest MAPE on the validation sets. The search ranges of hyperparameters are described as follows: batch size was chosen from [16, 32, 64, 128], learning rate was selected from [$5\times10^{-4}$, $5\times10^{-3}$, $1\times10^{-3}$, dropout rate was chosen from [0, 0.05, 0.1], and embedding dimension was selected from [32, 64, 128, 256]. Additionally, for CyclePatch models, both the number of intra-cycle encoder layers and the number of inter-cycle encoder layers were tuned from 0 to 12. The number of CNN layers is tuned from 1 to 8.

\subsection{Implementation Details of Baselines}\label{AppendixImplementation}
In this subsection, we introduce how we implemented the baselines in detail. First, the cycling data are resampled into $\textbf{X}\in\mathbb{R}^{3\times30000}$ according to the process introduced in Section~\ref{SecExperimentalDetails}. After that, the three variables in each cycle are normalized as follows:
\begin{gather}
	\textbf{p}_i=\textbf{p}_i/Q_{nominal} \\
	\textbf{v}_i=\textbf{v}_i/max(\textbf{v}_i) \\
	\textbf{s}_i=\textbf{s}_i/Q_{nominal}
\end{gather}
where $\textbf{p}_i\in\mathbb{R}^{300}$, $\textbf{v}_i\in\mathbb{R}^{300}$, $\textbf{s}_i\in\mathbb{R}^{300}$ represent the capacity, voltage, and current records in the $i^{th}$ cycle. The $Q_{nominal}$ denotes the nominal capacity. Let $\overline{\textbf{X}}_i$ represent the normalized cycling data of the $i^{th}$ cycle. Then, Transformer encoder, LSTM, BiLSTM, GRU, BiGRU can directly model $\overline{\textbf{X}}_i$ and produce prediction as follows:
\begin{gather}
	\textbf{H}=f(\overline{\textbf{X}}_{1:100})\\
	\hat{y}={\rm Projection}(\textbf{H})
\end{gather}
where $f$ denotes the processes described in the original papers, and $\overline{\textbf{X}}_{1:100}$ uses zeros as the placeholder when some cycles are not usable. The $\rm Projection(\textbf{H})$ flattens $\textbf{H}$ and then produces the prediction using a linear layer.

For the MLP, we first use a linear layer to embed the $\overline{\textbf{X}}_{1:100}$:
\begin{equation}
	\textbf{H}=\overline{\textbf{X}}_{1:100}\textbf{W}+\textbf{b}
\end{equation}
where $\textbf{W}\in\mathbb{R}^{30000\times D}$ and $\textbf{b}\in\mathbb{R}^D$ are learnable parameters. After that, stacked fully connected neural networks are employed to model $\textbf{H}$. In the $l^{th}$ stack, the process is:
\begin{gather}
	\hat{\textbf{z}}^l = \textbf{W}_2^l \sigma(\textbf{W}_1^l \textbf{z}^{l-1} + \textbf{b}_1^l) + \textbf{b}_2^l \\
	\textbf{z}^l = LN\left(\hat{\textbf{z}}^l + \textbf{z}^{l-1}\right)
\end{gather}
where $\textbf{z}^{l-1}\in\mathbb{R}^{D}$ is the input to the $l^{th}$ stack, and $\textbf{z}^{0}=\textbf{H}$. Let $L$ denote the number of stacks, MLP finally produces the prediction as follows:
\begin{gather}
	\hat{y}=\textbf{W}_{o}\textbf{z}^L+\textbf{b}_o
\end{gather}
where $\textbf{W}_o\in\mathbb{R}^{1\times D}$ and $\textbf{b}_o\in\mathbb{R}^{1}$ are learnable parameters.

Regarding the CNN, we first reshape $\overline{\textbf{X}}_{1:100}$ into the shape of $\textbf{S}\in\mathbb{R}^{3,100,300}$, where 3 indicates the three variables (voltage, current and capacity), 100 is the number of maximum input cycles, 300 denotes the resampled length. Then, a CNN is utilized to model the $\textbf{S}$. In each layer, a Conv2d is first used, followed by ReLU activation. After the activation, average pooling is used to extract useful information. After a preset number of layers, a linear projection is used to produce the prediction.

As for the DLinear, PatchTST, Autoformer, iTransformer, and MICN, we use the implementation in \url{https://github.com/thuml/Time-Series-Library}. Specifically, we slightly revise the classification implementations to make models usable for BLP.

\end{document}